\documentclass[sigconf]{acmart}

\AtBeginDocument{%
  }

\setcopyright{acmlicensed}
\copyrightyear{2018}
\acmYear{2018}
\acmDOI{XXXXXXX.XXXXXXX}

\acmConference[Conference acronym 'XX]{Make sure to enter the correct
  conference title from your rights confirmation emai}{June 03--05,
  2018}{Woodstock, NY}
\acmISBN{978-1-4503-XXXX-X/18/06}

\usepackage[utf8]{inputenc} 
\usepackage[T1]{fontenc}    
\usepackage{hyperref}       
\usepackage{url}            
\usepackage{booktabs}       
\usepackage{amsfonts}       
\usepackage{nicefrac}       
\usepackage{microtype}      
\usepackage{xcolor}         
\usepackage{amsmath}
\usepackage{algorithm}
\usepackage{enumitem}
\usepackage{picinpar}

\usepackage{graphicx}
\usepackage{algorithmicx}
\usepackage[noend]{algpseudocode}
\usepackage{subfigure}
\usepackage{multirow}
\usepackage{color}
\usepackage{balance}
\usepackage{enumitem}
\usepackage{hhline}
\usepackage[normalem]{ulem}
\usepackage{booktabs}
\usepackage{wrapfig}
\usepackage{cancel}
\usepackage{hyperref}
\usepackage{makecell}

\newcommand{\bL}{\ensuremath{\mathcal{L}}}

\newcommand{\bG}{\ensuremath{\mathcal{G}}}
\newcommand{\bD}{\ensuremath{\mathcal{D}}}

\newcommand{\bT}{\ensuremath{\mathcal{T}}}
\newcommand{\bX}{\ensuremath{\mathcal{X}}}

\renewcommand{\vec}[1]{\ensuremath{\mathbf{#1}}}

\newcommand{\stitle}[1]{\vspace{1mm} \noindent {\bf #1}}

\newcommand{\ie}{{\it i.e.}}

\newcommand{\method}[1]{\textsc{#1}}
\newcommand{\model}{\method{MDGPT}{}}
\newcommand{\cp}{mixing prompt{}}
\newcommand{\op}{unifying prompt{}}

\newcommand{\eat}[1]{}

\newcommand{\stkout}[1]{\ifmmode\text{\sout{\ensuremath{#1}}}\else\sout{#1}\fi}

\title{Text-Free Multi-domain Graph Pre-training:\\Toward Graph Foundation Models}

\def\equalcontribNew{%
      \ifnum\value{eqfn}=0%
        \footnote{Co-first authors with equal contribution. Part of the work was done while at Singapore Management University.}%
        \setcounter{eqfn}{\value{footnote}}%
      \else%
        \footnotemark[\value{eqfn}]%
      \fi%
    }%

\def\corresAuthor{%
      \ifnum\value{eqfn}=1%
        \footnote{Corresponding authors.}%
        \setcounter{eqfn}{\value{footnote}}%
      \else%
        \footnotemark[\value{eqfn}]%
      \fi%
    }%

\author{Xingtong Yu}
\affiliation{%
 \institution{Singapore Management University}
  \country{Singapore}}
\email{xingtongyu@smu.edu.sg}

\author{Chang Zhou}
\affiliation{%
 \institution{University of Science and Technology of China}
  \country{China}}
\email{chouchang21sy@mail.ustc.edu.cn,}

\author{Yuan Fang$^{\dagger}$}
\affiliation{%
  \institution{Singapore Management University}
  \country{Singapore}}
\email{yfang@smu.edu.sg}

\author{Xinming Zhang$^{\dagger}$}
\affiliation{%
  \institution{University of Science and Technology of China}
  \country{China}}
\email{inming@ustc.edu.cn}

\thanks{
    $^{\dagger}$Corresponding authors.
}
\begin{document}

\begin{abstract}
Given the ubiquity of graph data, it is intriguing to ask: Is it possible to train a graph foundation model on a broad range of graph data across diverse domains? A major hurdle toward this goal lies in the fact that graphs from different domains often exhibit profoundly divergent characteristics.
Although there have been some initial efforts in integrating multi-domain graphs for pre-training, they primarily rely on textual descriptions to align the graphs, limiting their application to text-attributed graphs.
Moreover, different source domains may conflict or interfere with each other, and their relevance to the target domain can vary significantly. 
To address these issues, we propose \model, a text free Multi-Domain Graph Pre-Training and adaptation framework designed to exploit multi-domain knowledge for graph learning. First, we propose a set of \emph{domain tokens} to to align features across source domains for synergistic pre-training.
Second, we propose a dual prompts, consisting of a \emph{unifying prompt} and a \emph{mixing prompt}, to further adapt the target domain with unified multi-domain knowledge and a tailored mixture of domain-specific knowledge. 
Finally, we conduct extensive experiments involving six public datasets to evaluate and analyze \model, which outperforms prior art by up to 37.9\%.(Codes are available at \url{https://anonymous.4open.science/r/MDGPT} for anonoymous review.)
\end{abstract}



\maketitle
\section{Introduction}\label{sec.intro}
Graph data are ubiquitous in many domains, owing to their ability to model complex relationships among entities. Thus, applications of graph data are widespread, ranging from e-commerce graphs for recommendation systems \cite{he2020lightgcn,wu2022graph}, to citation graphs for bibliographic study \cite{kanakia2019scalable,xiong2017explicit}, and social networks for user behavioral analysis \cite{fan2019graph,perozzi2014deepwalk}. These graphs constitute a vast knowledge repository with abundant and comprehensive information across various domains. Toward building a foundation model for graph analysis \cite{liu2023towards}, it is vital to train a universal graph model based on \textit{a broad range of  graph data from multiple domains}, which can be further adapted to solve \textit{diverse graph-centric tasks across different downstream fields}.

\begin{figure*}[t]
\centering
\includegraphics[width=0.9\linewidth]{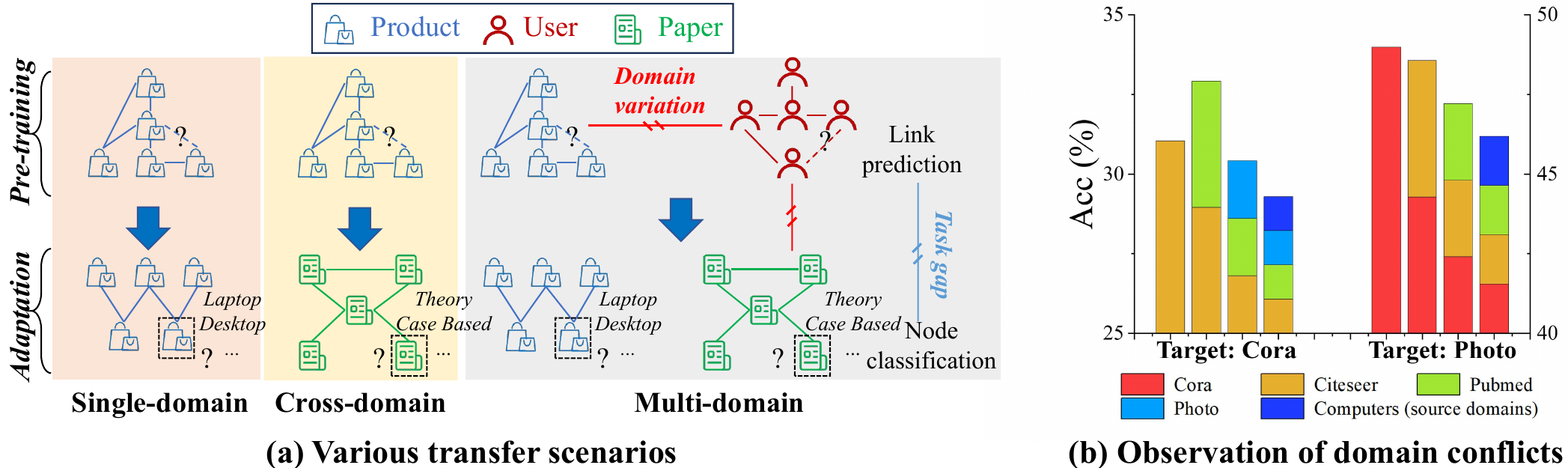}
\caption{Motivation of \model. (a) Different scenarios of pre-training and downstream adaptation. (b) Accuracy of one-shot node classification with DGI on two target domains, namely, \emph{Cora} and \emph{Photo}, as more source domains are added to pre-training.
}
\label{fig.intro-motivation}
\end{figure*}


However, mainstream graph learning methods fall short of achieving this goal. Conventional approaches often train graph neural networks (GNNs) \cite{kipf2016semi,velivckovic2017graph,yu2023learning} or graph transformers \cite{hu2020heterogeneous,yun2019graph,wu2024feasibility} in a supervised fashion, which requires not only re-training for each specific task, but also significant labeled data for each task.
Inspired by the advances in pre-trained models for language \cite{dong2019unified,kenton2019bert} and vision \cite{chen2021pre,bao2022beit} data, pre-training graph models \cite{hu2020gpt,velivckovic2018deep,you2020graph} has been widely explored to overcome the limitations of supervised approaches. They learn task-irrelevant, universal properties on unlabeled graphs, and the pre-trained graph model can be further tailored to different downstream tasks using some task-specific labels through finetuning \cite{velivckovic2018deep,you2020graph} 
To further narrow the \textit{task gap} between pre-training and downstream tasks, prompt learning methods \cite{liu2023graphprompt,sun2023all} emerge as a popular alternation of finetuning approaches. However, in prevailing solutions to graph pre-training, the pre-training and downstream graphs typically originate from the same dataset consisting of one or more graphs in the same domain, including different subgraphs of a large graph \cite{velivckovic2018deep,liu2023graphprompt}, or a collection of similar graphs \cite{hu2020gpt,qiu2020gcc}. 

Therefore, it is imperative to pre-train graph models on a broad spectrum of multi-domain graphs. 
Yet, graphs from different domains may exhibit profoundly divergent characteristics. For example, the small-world properties of a social network are not applicable to molecular graphs. 
The distinct nature of graphs from various domains poses a genuine hurdle in integrating broad multi-domain knowledge and achieving universal cross-domain transfer. 
On one hand, although some works have attempted cross-domain transfer from a single source domain \cite{ding2021cross,hassani2022cross,wang2021pre,wang2023cross,yang2019cross}, they fail to leverage comprehensive knowledge from a broad range of domains. On the other hand, a few recent studies \cite{liu2023one,tang2024higpt} employ large language models to process or extract node features based on associated textual descriptions in multi-domain graphs, where texts act as a unifying medium across different domains. Consequently, these approaches are limited to text-attributed graphs \cite{wen2023augmenting}, and do not extend to general graphs that lack explicit textual data.
In this work, we propose \model, a foundational graph framework for Text-Free \textbf{M}ulti-\textbf{D}omain \textbf{G}raph \textbf{P}re-\textbf{T}raining, to facilitate downstream adaptation to both seen and unseen domains. The solution is non-trivial due to two key challenges. 

First, \textit{how do we align multi-domain graphs in the pre-training phase?} Node features in various domains exhibit different distributions, often with unique dimensionality and semantics. For example, features in a citation network denote words in a paper, whereas in a commerce graph, they describe attributes of a product, as shown in Fig.~\ref{fig.intro-motivation}(a). Therefore, directly integrating multi-domain graphs may result in conflicts or interferences rather than synergy. As illustrated in Fig.~\ref{fig.intro-motivation}(b), adding more source domains often leads to worse performance for conventional graph pre-training methods such as DGI \cite{velivckovic2018deep}. Although some techniques like singular value decomposition (SVD) \cite{stewart1993early} can be used to align feature dimensions, the semantic meanings remain misaligned. A recent study \cite{zhao2024all} proposes a virtual node for each domain, linking all nodes within and then interlinking these virtual nodes. However, this method lacks an explicit mechanism for semantic alignment, leading to suboptimal performance. In \model, we propose a series of \textit{domain tokens} to align node features from different domains. Each domain-specific token is associated with a source domain, serving to bridge the semantic meanings in distinct domains. The token takes the form of a learnable vector, encouraging a synergistic multi-domain integration.

Second, \textit{how do we adapt multi-domain prior knowledge to downstream tasks in different domains?} The downstream task could be in either a seen or unseen domain. A recent multi-domain pre-training work \cite{zhao2024all} employs a finetuning or prompt-based approach for downstream adaptation. However, multi-domain prior knowledge contains not only unified global knowledge across domains, but also domain-specific knowledge in each domain. Conventional adaptation approaches for single-domain graph data fall short of effectively transferring such prior knowledge to the target domain.
In \model, we propose  \textit{dual prompts}, consisting of a \textit{\op} and a \textit{\cp}, to align the downstream target domain with the source domains, thus facilitating the transfer of multi-domain knowledge to downstream tasks. On one hand, \op\ is a learnable vector, serving as a bridge to holistically  align the target domain with the unified pre-trained knowledge from all source domains. On the other hand, \cp\ integrates the individual domain tokens, aligning the target domain with a tailored mixture of knowledge from each source domain, thereby obtaining domain-specific prior knowledge for more precise adaptation.


In summary, the contributions of this work are fourfold. 
(1) We propose \model, a text-free multi-domain pre-training and prompting framework on graphs, paving a plausible path towards graph foundation models.
(2) In pre-training, we design domain tokens to align nodes features, optimizing the pre-training model with multi-domain data.
(3) In downstream adaptation, we propose a dual-prompt design with a \op\ to align the target domain with the unified pre-trained prior, and a \cp\ for finer-grained domain-specific alignment. 
(4) We conduct extensive experiments on five benchmark datasets, demonstrating the superior performance of \model\ in comparison to the state-of-the-art approaches.

\section{Related Work}
Pre-training methods on graphs \cite{hu2020gpt,velivckovic2018deep,liu2023graphprompt,yu2024few,yu2023generalized} capitalize on the intrinsic properties of graph structures via self-supervised learning. Pre-trained knowledge can be transferred to downstream tasks through fine-tuning or parameter-efficient adaptation.  
However, these methods assume that the pre-training and downstream graphs come from the same domain (or very similar domains), such as different subgraphs of a large graph \cite{you2020graph,yu2023generalized}, or a collection of similar graphs in the same field \cite{hu2020gpt,qiu2020gcc}. As a result, these methods struggle to generalize across multi-domain graphs.

Another line of research on cross-domain learning addresses multiple domains \cite{ding2021cross,hassani2022cross,wang2021pre,wang2023cross}. These studies focus on transferring prior knowledge pre-trained on a single source domain to a different target domain by exploiting domain-invariant representations. However, they are only pre-trained on a single source domain, falling short of utilizing comprehensive multi-domain knowledge. Additionally, they are often tailored to specific tasks or domains \cite{ding2021cross,hassani2022cross,wang2021pre,wang2023cross}, or requiring domain expertise \cite{ding2021cross,wang2021pre}, which prevents them from generalize to diverse domains and tasks. 


Toward multi-domain graph pre-training, OFA \cite{liu2023one} and HiGPT \cite{tang2024higpt} employ a language model to derive and process node features from different domains through the medium of natural language, thereby limiting themselves to text-attributed graphs \cite{yu2024few,wen2023prompt}. For text-free graphs, GCOPE \cite{zhao2024all} introduces a series of domain-specific, interconnecting virtual nodes, which link to other nodes from each domain. While these virtual nodes facilitate some level of connection and alignment across the multiple domains, they do not explicitly align node features from different domains, and thus cannot effectively overcome domain conflicts. Moreover, it only transfers unified pre-trained knowledge to the target domain, overlooking source domain-specific knowledge. 
Inspired by multi-task pre-training \cite{wang2022multi,yu2023multigprompt}, in \model, we propose the notions of domain tokens and dual prompts to overcome the shortcomings of existing multi-domain graph pre-training methods. It is important to note that multi-task pre-training addresses the interference arising from various pre-training tasks within a single domain, which is distinct from our goal to mitigate domain conflicts involving multiple domains. 


\section{Preliminaries}
In this section, we present the related preliminaries and our problem definition.

\stitle{Graph.}
A graph is defined as \( G=(V,E,\vec{X}) \), where \( V \) is the set of nodes and \( E \) is the set of edges. \( \vec{X} \in \mathbb{R}^{|V| \times d} \) is the feature matrix for the nodes, so that the feature vector of node  \( v_i \) in \( V \) is \( \vec{x}_i \in \mathbb{R}^d \).
Furthermore, a set of graphs is denoted as  \( \mathcal{G} \). 

\stitle{Multi-domain pre-training.} \label{sec.pre.multi-domain}
Each pre-training graph belongs to a source domain \( D_{S_i} \), characterized by a node distribution $P^{(V)}_{S_i}$ and edge distribution $P^{(E)}_{S_i}$. 
Hence, given a set of multi-domain pre-training graphs $\bG_S$ and domains $\bD_S$, we have $\{(G_{1},D_{S_1}),(G_{2},D_{S_2}),\\\ldots,(G_{K},D_{S_K})\}$, where 
each graph $G_{i}\in \bG_S$ belongs to domain $D_{S_i}\in \bD_S$.
Note that more than one graph may belong to the same domain, i.e., we may have $S_i=S_j$ for some $i,j\le K$. 
In general, different domains follow different node and edge distributions, with varying node feature dimensions and semantics.

\stitle{Multi-domain few-shot adaptation.}\label{sec.down.cross-domain}
Given a downstream task, there is a set of graphs $\bG_T$ belonging to a target domain $D_T$.
The target domain is either \emph{seen} during pre-training (i.e., $\exists i \; D_T=D_{S_i}$), or \emph{unseen} (i.e., $\forall i\; D_T\ne D_{S_i}$).
We address few-shot node and graph classification tasks downstream, where each task only requires a small number of task-specific labels.
Specifically, for node classification within a graph \( G = (V, E, \vec{X}) \in \bG_T\), each node \( v \in V \) is associated with a class \( y \in Y \), where $Y$ is the set of node classes for the task. 
For graph classification across a series of graphs \( \bG_T \), each graph \( G \in \bG_T \) is associated with a class \( y \in Y \), where $Y$ denotes the set of graph classes.
In both node and graph classification, each class has only \( m \) labeled examples, where \( m \) is a small number. We call such a task as \( m \)-shot node or graph classification. 

\section{Proposed Approach}

In this section, we present our proposed model, \model, starting with an overview, followed by the pre-training and adaptation processes.

\begin{figure*}[t]
\centering
\includegraphics[width=0.85\linewidth]{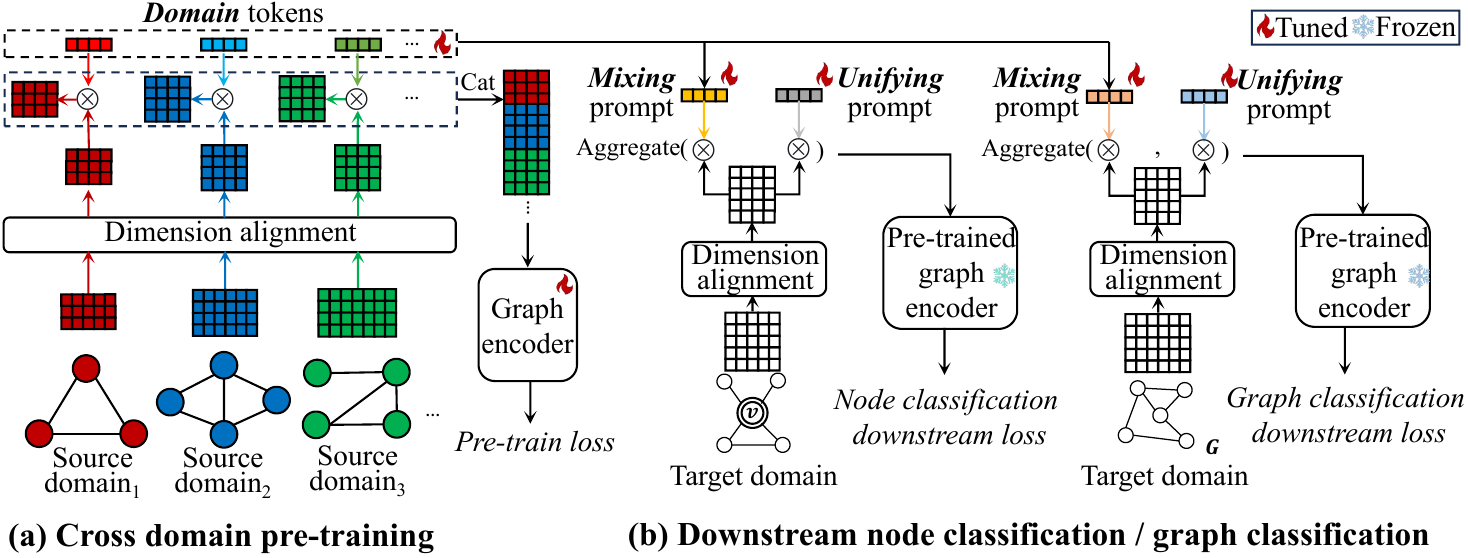}
\caption{Overall framework of \model.}
\label{fig.framework}
\end{figure*}

\subsection{Overall framework}
We present the overall framework of \model\ in Fig.~\ref{fig.framework}.

\stitle{Pre-training.} As shown in Fig.~\ref{fig.framework}(a), we first align the feature dimensions of graphs from multiple source domains. Next, we propose a series of \textit{domain tokens} to further unify the semantic spaces of features from different domains. Subsequently, we pre-train a graph encoder using a self-supervised task based on a universal task template following previous work \cite{liu2023graphprompt}. 

\stitle{Downstream adaptation.} As depicted in Fig.~\ref{fig.framework}(b), after an initial dimension alignment for the target domain, we propose \textit{dual prompts}. On one hand, we employ a \emph{\op}, a learnable vector designed to holistically align the target domain  with the unified pre-trained knowledge from all domains. On the other hand, we utilize a \emph{\cp}, a learnable aggregation of domain tokens to align the target domain with a tailored mixture of source domain-specific knowledge for fine-grained adaptation. The dual prompts then modify the downstream features to feed into the pre-trained and frozen graph encoder, adapting the downstream task solely through lightweight prompting.

\subsection{Multi-domain pre-training}
We first introduce the pre-training stage involving multiple distinct source domains. As detailed in Sect.~\ref{sec.pre.multi-domain}, node features across these domains exhibit unique distributions with varying feature dimensionality and semantic meanings, motivating us to align the dimensions and unify the semantics.

\stitle{Dimension alignment.}
The mismatch in node feature dimensions presents the first obstacle for jointly pre-training multi-domain graphs. A straightforward solution is to use a dimension alignment method to transform the features into consistent dimensions across domains. Specifically, given a graph $G_{i}=(V_{i}, E_{i}, \vec{X}_{i})$ from the source domain $D_{S_i}$, we transform its feature matrix as follows.
\begin{align}\label{eq.pre-train.dim-align}
    \tilde{\Vec{X}}_{i}=\mathtt{DA}_{S_i}(\Vec{X}_{i}),
\end{align}
where $\mathtt{DA}_{S_i}\colon\mathbb{R}^{|V| \times d_{S_i}}\rightarrow \mathbb{R}^{|V| \times \tilde{d}}$ is the dimension alignment function specific to domain $D_{S_i}$. Here $d_{S_i}$ is the feature dimension in domain $D_{S_i}$, and $\tilde{d}$ is the aligned dimension that is uniform across all domains.
Various choices for the alignment function exit, such as singular value decomposition (SVD) \cite{stewart1993early}, or a learnable multi-layer perceptron (MLP) \cite{kruse2022multi}. In our work, each alignment function $\mathtt{DA}_{S_i}$ is implemented using SVD. However, dimension alignment only transform the feature spaces into the same dimensions, yet they still occupy distinct semantic spaces across different domains. Direct pre-training on such dimension-aligned feature matrices from multiple domains would result in interference rather than synergy. 

\stitle{Semantic unification.}
To  unify the semantics of dimension-aligned features from different domains, we propose the notion of \emph{domain tokens}.
We inject a domain token $\Vec{t}_{S_i}$ into each source domain $D_{S_i}$, leading to a series of domain tokens, 
$\bT=\{\Vec{t}_{S_i}: \forall S_i \text{ s.t. } D_{S_i} \in \bD_{S}\}$. 
Each domain token adjusts the dimension-aligned features of graphs in the corresponding domain. 
Specifically, given a graph $G_{i}$ from domain $S_i$,  we further unify the semantics of its dimension-aligned features, $\tilde{\vec{X}}_{i}$, as follows.
\begin{align}
    \hat{\Vec{X}}_{i}=\Vec{t}_{S_i}\odot \tilde{\Vec{X}}_{i},
\end{align}
where $\hat{\Vec{X}}_{i}$ is the unified feature matrix, 
the domain token $\vec{t}_{S_i}$ is a $\tilde{d}$-dimensional learnable vector, and $\odot$ denotes element-wise multiplication. The unified feature spaces enable a synergistic integration of multi-domain graphs, facilitating the pre-trained model to extract a unified prior along with domain-specific knowledge from a broad range of source domains. Subsequently, given a set of source domain graphs $\bG_S$ and their  unified features $\bX_S=\{\hat{\Vec{X}}_{i}: \forall i \text{ s.t. } G_{i} \in \bG_{S}\}$, we can pre-train a graph encoder using a graph neural network or transformer backbone, as follows.
\begin{align}
    \Vec{H}_S=\mathtt{GE}(\bG_S,\bX_S;\Theta),
\end{align}
where $\mathtt{GE}$ represents graph encoder, and each row of $\Vec{H}_S$ corresponds to the output embedding of a node found in the collection of source domain graphs $\bG_S$. More precisely, let $\Vec{h}_{S,v}$  denote the embedding vector of node $v$, which is a node in one of the source domain graphs.

\stitle{Pre-training loss.}
Following previous work \cite{liu2023graphprompt,sun2022gppt}, we employ link prediction as the pre-training task, leveraging the abundant links in multi-domain graphs as self-supervision. In particular, we adopt the universal task template based on subgraph similarity \cite{liu2023graphprompt}, which enables compatibility among common graph-based tasks, including link prediction, node classification and graph classification. The compatibility of the pre-training task would facilitate adaptation to different downstream tasks.  

Specifically, the pre-training data consist of triplets taking the form $(v, a, B)$, where  $v,a$ are nodes from the source domain graphs in $\bG_S$, and $B=\{b_1,\ldots,b_m\}$ represents a  set of nodes from $\bG_S$. Here, $(v, a)$ is an edge present in $\bG_S$, while $B$ serves as the negative examples such that $(v, b_i)$ is not an edge, $\forall b_i \in B$. Such triplets can be readily sampled from $\bG_S$ to generate the pre-training dataset $\Omega_\text{pre}$.
Subsequently, utilizing the universal task template \cite{liu2023graphprompt}, we define the pre-training loss as
{\small\begin{align} \label{eq.pre-train-loss}
\textstyle
  \bL_{\text{pre}}(\Omega_\text{pre};\Theta,\bT)=-\sum_{(v,a,B)\in\Omega_\text{pre}}\ln\frac{\exp\left(\frac{1}{\tau}\mathtt{sim}(\vec{h}_{S,v},\vec{h}_{S,a})\right)}{\sum_{b_i\in B}\exp\left(\frac{1}{\tau}\mathtt{sim}(\vec{h}_{S,v},\vec{h}_{S,b_i})\right)},
\end{align}}%
where $\tau$ is a temperature hyperparameter, $\Theta$ denotes the parameters of $\mathtt{GraphEncoder}$, $\bT$ are the learnable domain tokens, 
and $\mathtt{sim}$ is a similarity function and is implemented as the cosine similarity following prior work.
The pre-training stage yields the pre-trained weights along with the  domain tokens, \ie, $(\Theta_{\text{pre}},\bT_{\text{pre}})=\arg\min_{\Theta,\bT} \mathcal{L}_{\text{pre}}(\Omega_\text{pre};\Theta,\bT)$, 
which are subsequently utilized in downstream tasks.

\subsection{Downstream adaptation}

In addition to addressing the task gap between pre-training and downstream tasks \cite{liu2023graphprompt,sun2022gppt}, we focus on transferring pre-trained multi-domain knowledge to downstream tasks, aiming to overcome the domain gap. Likewise, the downstream target domain, $D_T$, may have different feature dimensions from the aligned dimensions of the source domains.  
Therefore, the first step is to apply the same dimension alignment method as used in the pre-training phase. Given a downstream graph $G=(V,E,\Vec{X}) \in \bG_T$ from target domain $D_T$, 
we transform its feature matrix into
    $\tilde{\Vec{X}}=\mathtt{DA}_T(\Vec{X})$.

Afterwards, we propose the notion of \emph{dual prompts}, including \op\ and \cp, to further align the target domain with multiple source domains. One one hand, the \op\ aims to leverage the unified pre-trained knowledge from all source domains holistically. On the other hand, the \cp\ aims to leverage a tailored mixture of source domain-specific knowledge for finer-grained adaptation. We elaborate them below.

\stitle{Unifying prompt.}
Recall that during pre-training, we align node features across multiple source domains, integrating these domains into a unified semantic space for pre-training. To adapt the unified multi-domain knowledge to the downstream task, we propose a \op\ to align the target domain $D_T$ with the model pre-trained on the source domains. Specifically, we employ a learnable vector $\Vec{p}_\text{uni} \in \mathbb{R}^{\tilde{d}}$ as the unifying prompt, which can be used to modify the downstream features to achieve adaptation. 

\stitle{Mixing prompt.}
Compared to the unifying prompt for the unified multi-domain knowledge, the mixing prompt prioritizes knowledge specific to certain source domains. For the target domain, a closely related source domain is likely to provide more relevant prior knowledge.
Consequently, it is crucial to align the target domain with each source domain to a varying extent, prioritizing the most relevant one. 
Thus, we define the \cp, $\Vec{p}_\text{mix} \in \mathbb{R}^{\tilde{d}}$, as a learnable aggregation of pre-trained domain tokens $\bT_\text{pre}$, where each domain token $\vec{t}_{S_i}\in\bT_\text{pre}$ can be deemed as a pre-trained ``prompt'' specific to its corresponding source domain $D_{S_i}$. Concretely, let
\begin{align}\textstyle
\Vec{p}_\text{mix}=\sum_{i=1}^K \gamma_i\Vec{t}_{S_i},
\end{align}
where $\Gamma=\{\gamma_1,\ldots,\gamma_K\}$ are the learnable mixing coefficients for the source domains. 
The mixing prompt then modifies the  downstream features to achieve source domain-specific  adaptation.

\stitle{Prompt tuning.}
As discussed, the dual prompts are used to modify the downstream features,  $\tilde{\Vec{X}}$, to fully leverage the multi-domain pre-trained knowledge. 
More concretely, consider a downstream graph $G \in \bG_T$ from the target domain $D_T$.
We apply the unifying prompt and mixing prompt separately to its dimension-aligned feature matrix, $\tilde{\Vec{X}}$, before feeding them to the pre-trained graph encoder to obtain the final node representations. 
\begin{align}
\scriptsize
    \Vec{H}=\mathtt{GE}(G,\vec{p}_\text{uni}\odot\tilde{\Vec{X}};\Theta_{\text{pre}})+\mathtt{GE}(G,\vec{p}_\text{mix}\odot \tilde{\Vec{X}};\Theta_{\text{pre}}),
\end{align}
where $\Theta_{\text{pre}}$ denotes the pre-trained weights that are kept frozen during downstream adaptation. The final embedding matrix, $\vec{H}$, results from a combination of both unified and a tailored mixture of pre-trained knowledge and will be used to compute the downstream task loss  $\bL_{\text{down}}(\Vec{H};\Vec{p}_\text{uni},\Gamma)$. To optimize the downstream loss, both the unifying prompt $\Vec{p}_\text{uni}$ and the mixing coefficients $\Gamma$ for the mixing prompt are tunable, whereby the pre-trained weights $\Theta_0$ and domain tokens $\bT_0$ remain  frozen. Hence, only a small number of parameters are updated downstream, offering a more parameter-efficient approach than conventional fine-tuning. Hence, our approach \model\ is especially amenable to few-shot settings, where the downstream task is provided with only a limited number of labeled examples. 

In general, for downstream node or graph classification tasks, $\bL_\text{down}$ follows the universal task template based on subgraph similarity \cite{liu2023graphprompt}, similar to the pre-training loss $\bL_\text{pre}$.
Consider a labeled training set $\Omega_\text{down}=\{(x_1,y_1),(x_2,y_2),\ldots\}$, where $x_i$ is either a node or a graph, and $y_i\in Y$ is the class label of $x_i$ among a set of classes $Y$. The loss is then defined as
{\small\begin{align}
\textstyle
    \bL_{\text{down}}(\Vec{H};\Vec{p}_\text{uni},\Gamma)=-\sum_{(x_i,y_i)\in \Omega_\text{down}}\ln\frac{\exp\left(\frac{1}{\tau}\text{sim}(\vec{{h}}_{x_i},\overline{\vec{{h}}}_{y_i})\right)}{\sum_{y\in Y}\exp\left(\frac{1}{\tau}\text{sim}(\vec{{h}}_{x_i},\overline{\vec{{h}}}_{y})\right)},
\end{align}}%
where $\vec{{h}}_{x_i}$ denotes the final embedding of node/graph $x_i$, and $\overline{\vec{{h}}}_y$ is the prototype embedding of class $y$, representing the mean of the training instances of class $y$. For a node instance $v$, $\vec{{h}}_{v}$ is a row of $\Vec{{H}}$. For a graph instance $G$, 
$\vec{{h}}_{G}$ is a readout of the final embedding matrix $\Vec{H}$. 

\section{Experiments}
In this section, we conduct experiments to evaluate \model, and analyze the empirical results.

\subsection{Experimental Setup}
\stitle{Datasets.}
We conduct experiments on five benchmark datasets.
(1) \emph{Cora} \cite{mccallum2000automating}, (2) \emph{Citeseer} \cite{sen2008collective} and (3) \emph{Pubmed} \cite{sen2008collective} are citation graphs from different fields (e.g., computing and biomedical sciences). Each dataset consists of a single graph. Each dataset consists of a single graph, with nodes representing scientific publications and edges denoting citations. In line with previous work \cite{kipf2016semi,velivckovic2017graph}, we treat the edges as undirected.
(4) \emph{Photo} \cite{shchur2018pitfalls} and (5) \emph{Computers} \cite{mcauley2015image} are both e-commerce networks from Amazon in different categories (e.g., photography and computer related products). Each dataset comprises a single graph, where nodes represent products and edges signify frequent co-purchases between products.
(6) \textit{Reddit} \cite{hamilton2017inductive} is a social network, where nodes are post and edges represent interactions such as comments and replies between posts. Note that different citation or e-commerce graphs have distinct features. We 
present additional details of datasets in Appendix~\ref{app.dataset}.


\stitle{Setup of multi-domain pre-training.}
To evaluate the performance of multi-domain pre-training, we focus on the more challenging scenario of generalizing to unseen domains in downstream tasks. For \textit{Cora}, \textit{Citesser}, \textit{Pubmed}, \textit{Photos} and \textit{Computers}, we designate each dataset individually as the downstream target domain, while employing the remaining four datasets as the four source domains during multi-domain pre-training. When \textit{Reddit} is used as the target domain, we leverage all five of the previous datasets as source domains. 
Details of multi-domain transfer to unseen and seen domains are provided in Appendix~\ref{app.setup} and Sect.~\ref{app.exp.seen}, respectively.

\stitle{Setup of downstream tasks.}
We conduct two types of downstream task, namely, \textit{node classification}, 
and \textit{graph classification}. We set the tasks in an $m$-shot classification, i.e., for each class, we randomly sample $m$ instances (nodes or graphs) as supervision. 
Note that each dataset only comprises a single graph, which cannot be directly applied to graph classification. Therefore, following previous work \cite{lu2021learning,yu2023hgprompt}, we generate a set of graphs by constructing ego-networks centered on the target nodes (i.e., those with class labels) in each dataset. Subsequently, we conduct graph classification across these ego-networks, each is labeled identically to its corresponding central node.
Given that the $m$-shot tasks are balanced classification, we evaluate performance using accuracy, consistent with prior studies \cite{yu2023generalized,wang2020graph,liu2021relative}. 
Note that we pre-train the graph encoder once for each dataset and then apply the same pre-trained model for all types of downstream tasks.

\begin{table*}[tbp] 
    \centering
    \small
     \addtolength{\tabcolsep}{1mm}
    \caption{Accuracy of one-shot node classification, where each column indicates a target domain. 
    }
    \label{table.node-classification}%
    \resizebox{0.9\linewidth}{!}{%
    \begin{tabular}{l|c|c|c|c|c|c}
    \toprule
   {{\tiny Method} \textbackslash\ {\tiny Target domain}}   & Cora & Citeseer & Pubmed & Photo & Computers & Reddit
      \\\midrule\midrule
    \method{GCN} 
    & 28.57 $\pm$ \phantom{0}5.07 & 26.82 $\pm$ \phantom{0}5.92  & 40.03 $\pm$ \phantom{0}8.53  & 46.37 
 $\pm$ 10.58  & 34.77   $\pm$ 11.76 & 60.12  $\pm$ \phantom{0}5.86 
    
\\ 
    \method{GAT} 
    & 28.40 $\pm$ \phantom{0}6.25 & 23.79  $\pm$ \phantom{0}2.96  & 38.99  $\pm$ \phantom{0}4.95   & 29.42 
 $\pm$ \phantom{0}7.96 & 31.57 $\pm$ \phantom{0}5.87 & 47.79 $\pm$ \phantom{0}6.73    
\\\midrule
    \method{DGI}
    & 29.30 $\pm$ \phantom{0}5.82  & 30.03 $\pm$ \phantom{0}4.88  
    & 41.85 $\pm$ \phantom{0}7.78  & 46.18 $\pm$ \phantom{0}7.48 & 39.37 
 $\pm$ \phantom{0}8.07  & 60.48  $\pm$ \phantom{0}7.38 
\\
    \method{GraphCL}
    & 34.94 $\pm$ \phantom{0}6.49  & 30.58 $\pm$ \phantom{0}4.58  
    & 40.36 $\pm$ \phantom{0}7.81  & 42.26 $\pm$ \phantom{0}7.31  & \underline{45.28} $\pm$ \phantom{0}6.59 & 59.80 $\pm$ \phantom{0}4.96  
\\\midrule
    Hassani 
        & 25.75 $\pm$ \phantom{0}6.52  & 31.67 $\pm$ \phantom{0}6.35  
    & 42.18 $\pm$ \phantom{0}8.46  & 47.00   $\pm$ \phantom{0}8.52 & 38.14 $\pm$ \phantom{0}7.91 
  & 52.40   $\pm$ \phantom{0}4.88 
    \\\midrule
    \method{GPPT}
    & 15.37 $\pm$ \phantom{0}4.51	& 23.24  $\pm$ \phantom{0}2.94  & 36.56  $\pm$ \phantom{0}5.31  & 16.19  
 $\pm$ \phantom{0}4.73 & 19.22  $\pm$ \phantom{0}8.71 & 52.10  $\pm$ \phantom{0}7.42 
\\
    \method{GraphPrompt}
    & 35.90 $\pm$ \phantom{0}7.10  & 32.76 $\pm$ \phantom{0}7.66  
    & 43.34 $\pm$ 10.66 & \underline{49.88} $\pm$ \phantom{0}8.31 & 43.03 $\pm$ 10.35  & \underline{65.49} $\pm$ \phantom{0}6.52 
\\
    \method{GPF}
    & \underline{37.84} $\pm$ 11.07  & \underline{37.61} $\pm$ \phantom{0}8.87    
    & \underline{46.36} $\pm$ \phantom{0}7.48 & 49.42  $\pm$ \phantom{0}7.04  &37.00  $\pm$ \phantom{0}6.52  & 64.02 $\pm$ \phantom{0}7.98 

\\\midrule

    \method{GCOPE}
    & 33.38 $\pm$ \phantom{0}6.86  & 35.56 $\pm$ \phantom{0}6.81  
    & 42.10 $\pm$ \phantom{0}8.07 & 48.52  $\pm$ \phantom{0}7.78 & 40.22 
  $\pm$ \phantom{0}7.82 & 61.35
  $\pm$ \phantom{0}5.30  	\\
  
      \method{\model}
    & \textbf{42.26} $\pm$ 10.18 & \textbf{42.40} $\pm$ \phantom{0}9.26 
    & \textbf{49.82} $\pm$ \phantom{0}8.38 & \textbf{64.82} $\pm$ 10.53 & \textbf{49.77} $\pm$ 11.00 & \textbf{67.80} $\pm$ \phantom{0}6.84  
\\    \bottomrule
        \end{tabular}}
        \\
   \parbox{0.9\linewidth}{Results are reported in percent. The best method is bolded and the runner-up is underlined. Table~\ref{table.graph-classification} follows the same style. }
\end{table*}

\begin{table*}[tbp] 
    \centering
    \small
     \addtolength{\tabcolsep}{1mm}
    \caption{Accuracy of one-shot graph classification, where each column indicates a target domain. 
    }
    \label{table.graph-classification}%
    \resizebox{0.9\linewidth}{!}{%
    \begin{tabular}{l|c|c|c|c|c|c}
    \toprule
   {\tiny Method} \textbackslash\ {\tiny Target domain}   & Cora & Citeseer & Pubmed & Photo & Computers & Reddit
      \\\midrule\midrule
    \method{GCN} 
    & 39.74  $\pm$ 13.70  & 25.58 $\pm$ \phantom{0}9.13 & 50.50  $\pm$ 10.85  & 56.97 
 $\pm$ 10.53  & 39.48 $\pm$ 12.19 & 58.63 $\pm$ \phantom{0}5.96

\\ 
    \method{GAT} 
    & 26.73 $\pm$ \phantom{0}9.06 & 25.33  $\pm$ \phantom{0}6.01 & 38.06  $\pm$ \phantom{0}7.14   & 47.75 
 $\pm$ 11.58 & 36.59  $\pm$ 12.55  & 66.10   $\pm$ \phantom{0}7.71 

\\\midrule
    \method{InfoGraph}
    & 40.23 $\pm$ 10.00 & 30.28 $\pm$ \phantom{0}6.70  
    & 49.27 $\pm$ \phantom{0}9.75 & 59.44  $\pm$ 10.25 & 40.58  $\pm$ 10.30 & 64.19 $\pm$ \phantom{0}6.88

\\
    \method{GraphCL}
    &  41.46  $\pm$  10.09  & 32.93  $\pm$ \phantom{0}8.04 
    & 49.49 $\pm$ \phantom{0}9.14  & 57.58  $\pm$ 10.45  & 42.68  $\pm$ 10.55 & 63.27 $\pm$ \phantom{0}5.80 
\\\midrule
    Hassani 
    & 40.77 $\pm$ \phantom{0}8.04  & 31.85  $\pm$ \phantom{0}6.22 
    & 49.79 $\pm$ \phantom{0}9.96  & 60.29  $\pm$ 10.77 & 37.65 $\pm$ 10.18 
  & 65.64 $\pm$ \phantom{0}6.65  
    \\\midrule
    \method{GraphPrompt}
    & 41.59 $\pm$ \phantom{0}9.85 & \underline{38.92} $\pm$ \phantom{0}8.97  
    & \underline{51.42} $\pm$ 11.15 & \underline{60.86} $\pm$ \phantom{0}9.28 & \underline{45.93} $\pm$ 11.70  & 64.61 $\pm$ \phantom{0}6.39 
    \\
    \method{GPF}
    & 41.46 $\pm$ 10.09  & 38.87 $\pm$ \phantom{0}9.22  
    & 48.07 $\pm$ 10.52  & 60.76  $\pm$ \phantom{0}9.56  &45.65  $\pm$ 10.31  & 65.91  $\pm$ \phantom{0}6.28 
 
\\\midrule

    \method{GCOPE}
    & \underline{42.61} $\pm$ \phantom{0}8.57  & 35.27 $\pm$ \phantom{0}8.48  
    & 49.42 $\pm$ \phantom{0}9.52  & 58.50 $\pm$ \phantom{0}9.70 	& 45.00 $\pm$ 10.30 & \underline{67.49} $\pm$ \phantom{0}7.20 
    \\
    \method{\model}
    & \textbf{48.36} $\pm$ 11.34 & \textbf{44.28} $\pm$ 10.16 
    & \textbf{54.34} $\pm$ \phantom{0}9.76  & \textbf{64.08} $\pm$ \phantom{0}9.85 & \textbf{48.29} $\pm$ 11.35  & \textbf{68.92} $\pm$ \phantom{0}7.32
\\    \bottomrule
        \end{tabular}}
\end{table*}

\stitle{Baselines.}
We evaluate the performance of \model\ against state-of-the-art approaches across five primary categories as outlined below. 
(1) \emph{End-to-end graph neural networks}: GCN \cite{kipf2016semi} and GAT \cite{velivckovic2017graph} utilize neighborhood aggregation to iteratively gather messages from adjacent nodes, training in an end-to-end manner without pre-training.
(2) \emph{Graph pre-training models}: DGI/InfoGraph\footnote{Original DGI only operates at the node level, while InfoGraph extends it to the graph level. In our experiments, we apply DGI to node classification, and InfoGraph to graph classification.} \cite{velivckovic2018deep,sun2019infograph} and GraphCL \cite{you2020graph}. follow the ``pre-train, fine-tune'' paradigm. Specifically, they first pre-train the GNN to exploit the intrinsic attributes of the graphs and then fine-tune the pre-trained model on downstream tasks according to task-specific labels.
(3) \textit{Graph cross-domain model}: Hassani \cite{hassani2022cross} pre-trains a attention-based GNN from both contextual and topological views for cross-domain adaptation.
(4) \emph{Graph prompting models}: GPPT\footnote{GPPT is specifically designed for downstream node classification task and is not suitable for graph classification. Therefore, in our experiments, we exclusively use GPPT for node classification.} \cite{sun2022gppt}, GPF\cite{fang2022universal} and GraphPrompt \cite{liu2023graphprompt} fall under this category. They employ a self-supervised pre-training tasks and unify downstream tasks into a general template as the pretext task. Subsequently, a single type of prompt is tuned for downstream adaptation.
(5) \emph{Multi-domain pre-training models}: GCOPE \cite{zhao2024all} is pre-trained on multi-domain datasets through self-supervised tasks, followed by fine-tuning or prompting for downstream adaptation.

We provide further descriptions of these baselines in Appendix~\ref{app.baselines}, and their implementation and configuration details, along with those of our approach \model, in Appendix~\ref{app.parameters}.

\subsection{Few-shot performance evaluation}\label{sec.exp.per}
We first evaluate one-shot classification tasks. Subsequently, we vary the number of shots to investigate their impact on performance.

\stitle{One-shot performance.}\label{exp.main}
We present the results of one-shot node and graph classification tasks on unseen domains in Tables~\ref{table.node-classification} and~\ref{table.graph-classification}, respectively. (Additional results on seen domains are presented in Sect.~\ref{app.exp.seen}.)
We make the follow observations.
(1) \model\ surpasses all baseline methods across all settings, outperforming the best competitor by 3.5--37.9\% on node classification and 4.3--13.9\% on graph classification.
The results demonstrate its effectiveness in multi-domain pre-training and in downstream adaptation to unseen domains. 
(2) Another text-free multi-domain pre-training method, GCOPE, significantly lags behind \model.
Its suboptimal performance can be attributed to two factors. First, it only performs implicit domain alignment via virtual nodes, without explicitly unifying feature spaces across domains. Second, it overlooks source domain-specific knowledge in downstream adaptation. The results also underscore the importance of our domain tokens in aligning diverse domains and enabling a tailored mixture of source domains.  
(3) GPPT is only comparable to, if not worse than, other baselines since it is not specifically designed for few-shot learning. 

\begin{figure}[t]
\centering 
\includegraphics[width=1\linewidth]{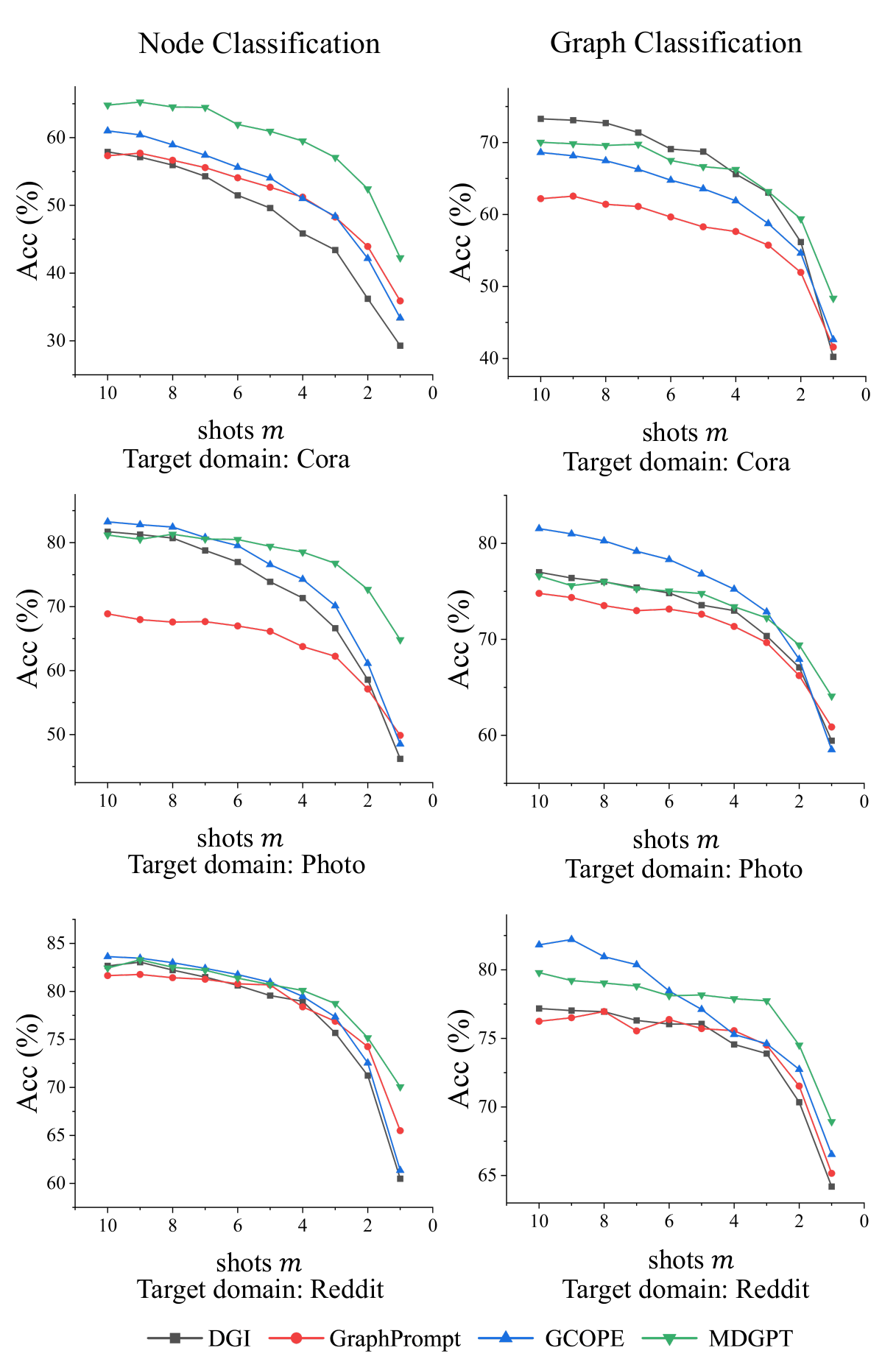}
\caption{Impact of number of shots on node classification on three target domains.}
\label{fig.fewshot}
\end{figure}

\stitle{Few-shot performance.}
To examine the performance of \model\ with more labeled data, we vary the number of shots in both node and graph classification tasks, and report the results in Fig.~\ref{fig.fewshot}.
Our findings are as follows.
(1) \model\ significantly surpasses all baselines in low-shot settings with very limited labeled data (e.g., $m \le 5$), highlighting the ideal use case for our approach. 
(2) As the number of shots increases, the performance of all methods improves as anticipated. However, \model\ continues to perform competitively, if not better.

\begin{table}[tb]
    \centering
    \small
    \addtolength{\tabcolsep}{-1mm}
    \caption{Analysis on homophilic and heterophilic data.}
    \label{table.homo-hetero}%
    \resizebox{0.85\linewidth}{!}{%
    \begin{tabular}{@{}l|l|cc@{}}
    \toprule
    \multirow{2}{*}{Target domain} & \multirow{2}{*}{Source domain} & \multicolumn{2}{c}{Accuracy (\%)} \\
    & & GCOPE & MDGPT  \\
    \midrule\midrule
    \multirow{3}{*}{Pubmed}
    & Cora & 40.92 & 45.59\\ 
    & Cora + Citeseer  & 40.58 & 46.72\\
    & Cora + Citeseer + Wisconsin & 43.14 & 48.51\\
    \midrule
    \multirow{3}{*}{Wisconsin}
    & Texas & 27.37  & 27.75\\
    & Texas + Cornell & 27.55 & 27.73\\
    & Texas + Cornell + Cora & 27.14 & 29.78\\
    \midrule
    \multirow{3}{*}{Chameleon} 
    & Wisconsin & 23.14 & 24.81\\
    & Wisconsin + Squirrel & 24.13 & 24.95\\
    & Wisconsin + Squirrel + Cora & 22.61 & 25.36\\
    \bottomrule
    \end{tabular}}
    
\end{table}

\begin{table*}[tb]
    \centering
    \small
    \caption{Model ablation study on the effects of  key components.}
    \label{table.ablation}%
    \begin{tabular}{@{}l|ccc|ccccc|ccccc@{}}
    \toprule
    \multirow{2}*{Methods}
    & Domain & Mixing & Unified &\multicolumn{5}{c|}{Target domain for node classification} &\multicolumn{5}{c}{Target domain for graph classification}\\
    &token &prompt &prompt & Cora & Citeseer & Pubmed & Photo & Computers & Cora & Citeseer & Pubmed & Photo & Computers\\
    \midrule\midrule
    \method{Variant 1}
    & $\times$ & $\times$ & $\times$ &32.11  & 32.45  & 37.47  & 49.69  & 38.98  & 43.12  & 33.36  & 52.89  & 56.04  & 40.79   \\ 
    \method{Variant 2}
    & $\times$ & $\times$ & $\checkmark$ &35.90  & 38.16  & 43.34  & 49.88  & 43.03  & 41.59  & 38.92  & 51.42  & 60.86  & 45.93   \\ 
    \method{Variant 3} 
    & $\checkmark$ & $\times$ & $\times$ & 36.50  & 40.05  & 47.02  & 50.64  & 43.37  & 43.39  & 39.18  & 53.07  & 57.57  & 43.38   \\  
    \method{Variant 4}
    & $\checkmark$ & $\checkmark$ & $\times$ &37.24  & 41.47  & 48.22  & 56.59  & 41.30  & 43.94  & 42.45  & 49.95  & 56.83  & 44.11   \\ 
    \method{\model}
    & $\checkmark$ & $\checkmark$ & $\checkmark$ & \textbf{42.26} & \textbf{42.40} & \textbf{49.82} & \textbf{64.82} & \textbf{49.77} & \textbf{48.36} & \textbf{44.28} & \textbf{54.34} & \textbf{64.08} & \textbf{48.29} \\
    \bottomrule
    \end{tabular}
\end{table*}


\subsection{Ablation Study}
To comprehensively analyze the performance of \model, we conduct two ablation studies as follows.

\begin{figure*}[t]
\centering
\begin{minipage}[b]{0.36\textwidth}
\centering
\includegraphics[width=0.8\linewidth]{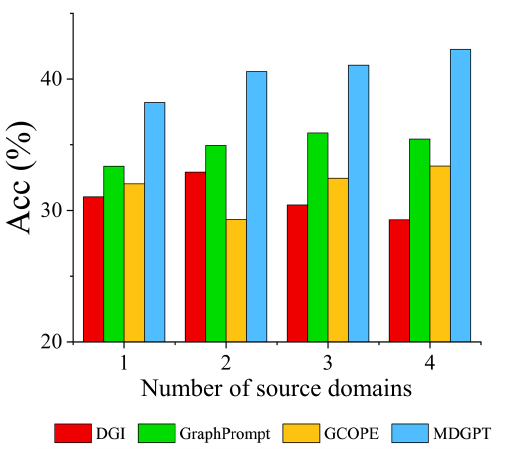}
\vspace{-2mm}
\caption{Data ablation study with a growing number of source domains.}\vspace{-2mm}
\label{fig.ablation}
\end{minipage}%
\hspace{5.4mm}%
\begin{minipage}[b]{0.6\textwidth}
\centering
\includegraphics[width=0.95\linewidth]{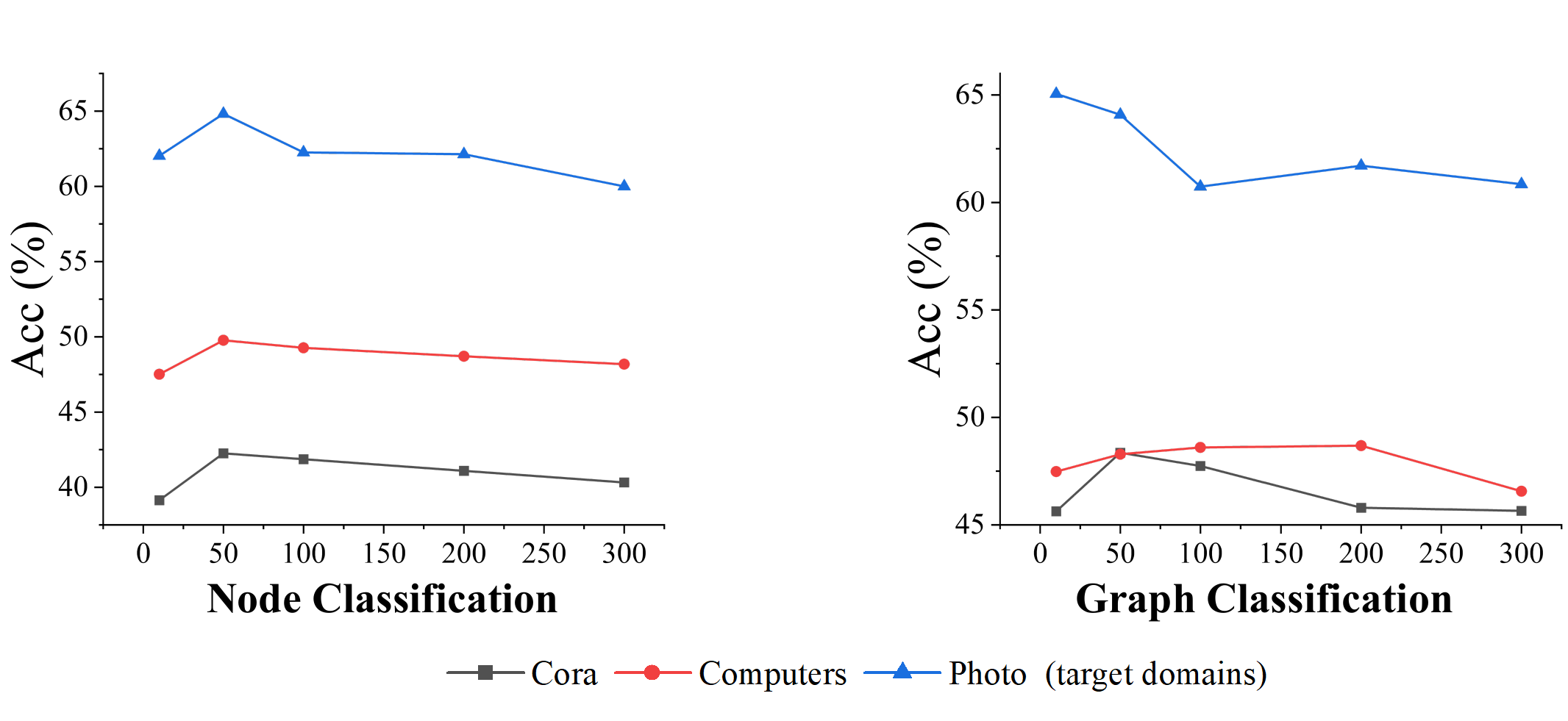}
\vspace{-5mm}
\caption{Sensitivity study of $\tilde{d}$, the aligned feature dimension across domains, on three target domains.}\vspace{-2mm}
\label{fig.para}
\end{minipage}
\end{figure*}

\stitle{Data ablation.} We evaluate the impact of utilizing more source domains, by iteratively adding \textit{Citeseer}, \textit{Pubmed}, \textit{Photo}, and \textit{Computers} to pre-training, in this order, while the target domain is fixed to \textit{Cora}.
As the number of source domains grows from one to four, we perform node classification on \textit{Cora} using several strong baselines and \model, as shown in Fig.~\ref{fig.ablation}. We observe that for all baselines, adding more datasets tends to cause domain conflicts. Specifically, DGI performs worse when \textit{Photo} and \textit{Computers} are added, GraphPrompt performs worse when \textit{Computers} is added, and GCOPE performs worse when \textit{PubMed} is added. In contrast, \model\ consistently performs better when more source domains are introduced, demonstrating the effectiveness of our multi-domain pre-training. 

\stitle{Model ablation.} We compare \model\ to its ablated variants obtained by excluding the use of domain tokens, mixing prompts, or unifying prompts. These variants and their corresponding results on downstream tasks are presented in Table~\ref{table.ablation}. 
The findings confirm the importance of each design element.
First, the use of domain tokens and mixing prompt is crucial. Notably, Variant 3, which includes domain tokens, significantly outperforms Variants 1 and 2, demonstrating the effectiveness of domain tokens in aligning multiple source domains. Furthermore, Variant 4 surpasses Variant 3, highlighting the additional benefit of using mixing prompts for downstream adaptation.
Second, omitting the unifying prompt results in reduced performance, as evidenced by the higher accuracy of Variant 2 compared to Variant 1.
Lastly, the dual prompts with both mixing and unifying prompts are beneficial, enabling \model\ to achieve superior performance.



\subsection{Hyperparameter Analysis}
We evaluate the impact of the feature dimension, $\tilde{d}$, after performing domain alignment in Eq.~\eqref{eq.pre-train.dim-align}. 
We vary $\tilde{d}$ and report the corresponding performance in Fig.~\ref{fig.para}.
We observe that, for both node and graph classification, as $\tilde{d}$ grows from a low value, the performance improves initially since higher dimensions increases model capacity. However, after reaching a peak (around $\tilde{d}=50$), accuracy starts to gradually decline as $\tilde{d}$ grows further, due to the amplified semantic gap in these feature dimensions across various domains.
Based on the above observation, $\tilde{d}=50$ is a robust setting in most datasets and tasks.


\subsection{Homophily Sensitivity}
To further evaluate the robustness of \model\ across homophilic and heterophilic graphs, we conduct cross-domain experiments on homophilic datasets (\textit{Cora}, \textit{Citeseer}, \textit{Pubmed}) and heterophilic datasets (\textit{Texas} \cite{pei2020geom}, \textit{Wisconsin} \cite{pei2020geom}, \textit{Cornell} \cite{pei2020geom}, \textit{Squirrel} \cite{rozemberczki2021multi}, \textit{Chameleon} \cite{rozemberczki2021multi}). The results are shown in Table~\ref{table.homo-hetero}. We observe that regardless of whether the source or target domain graphs are homophilic or heterophilic, \model\ effectively leverages multi-domain knowledge and bridges the gap between various domains and consistenely outperforms GCOPE. This further demonstrates the effectiveness of \model.

\subsection{Node Classification on Seen Domains}\label{app.exp.seen}
We further conduct one-shot node classification tasks on seen target domains. Specifically, we pre-train on the source domains including \textit{Cora},  \textit{Citeseer}, \textit{Pubmed}, \textit{Photo} and \textit{Computers}, and evaluate node classification on a seen target domain, \textit{Cora} or \textit{Citeseer}, separately. We illustrate the results of several competitive baselines and \model\ in Table~\ref{app.table.node-classification}, and observe the same trend as those observed in in Sect.~\ref{exp.main}, showing the robustness of \model\ in dealing with both seen and unseen domains.

\begin{table}[tbp] 
    \centering
    \small
    \caption{Accuracy of one-shot node classification on seen target domains. 
    }
    \label{app.table.node-classification}%
    \begin{tabular}{l|c|c}
    \toprule
   {Method} \textbackslash\ {Target domain}   & \multicolumn{1}{c|}{Cora} & \multicolumn{1}{c}{Citeseer}
      \\\midrule\midrule
    \method{DGI}
    & 33.57 $\pm$ \phantom{0}6.83 & 32.90 $\pm$ \phantom{0}6.10  
\\
    \method{GraphCL}
    & 38.73 $\pm$ \phantom{0}7.43  & 32.46 $\pm$ \phantom{0}4.75      
\\
    \method{GraphPrompt}
 & \underline{45.98} $\pm$ \phantom{0}9.43  & \underline{36.45} $\pm$ \phantom{0}8.38     
\\
    \method{GCOPE}
& 34.95 $\pm$ \phantom{0}7.45 & 35.18 $\pm$ \phantom{0}7.76
\\
      \method{\model}
      & \textbf{47.44} $\pm$ 10.03 & \textbf{44.96} $\pm$ 10.12
\\    \bottomrule
        \end{tabular}
        \\
\end{table}

\subsection{Parameters Efficiency}\label{app.exp.para-effi}
We evaluate the parameter efficiency of \model\ compared to representative baselines. We present the number of tunable parameters during downstream adaptation in Table~\ref{table.parameters-num}.
For GCN, as it follows an end-to-end supervised paradigm, all parameters in the model have to be updated, resulting in the most parameters.
For GraphCL and GCOPE, we update only the downstream classifier while freezing the pre-trained model, significantly reducing the number of tunable parameters.
For GraphPrompt, since it only tunes a light-wight prompt containing fewer parameters than a classifier, the number of parameters is further reduced.
Finally, \model\ is the most parameter-efficient adaptation, as only the dual prompts are updated. Specifically, the unifying prompt has the same dimension as the unified features, which is generally smaller than the original feature dimension, and the mixing prompt's dimension is same as the number of source domains, which is a quite small compared to the feature dimension. Despite having the fewest parameters, \model\ significantly outperforms all baselines, further demonstrating the effectiveness of our design.

\begin{table}[tbp] 
    \centering
    \small
    \caption{Comparison of the number of tunable parameters during the downstream adaptation stage. 
    }
    \label{table.parameters-num}%
    \begin{tabular}{l|cccc}
    \toprule
   { Method} \textbackslash\ { Target domain} 
   & \multicolumn{1}{c}{Cora} & \multicolumn{1}{c}{Citeseer}  & \multicolumn{1}{c}{Computers} & \multicolumn{1}{c}{Reddit}
    \\\midrule\midrule
    \method{GCN} 
    & 14,592 & 14,336
    &15,360 &23,296
\\     
    \method{GraphCL}
    & 1,792  & 1,536  
     & 2,560 &10,496\\
    \method{GraphPrompt}
    & 256  & 256  
    & 256 & 256 \\
    \method{GCOPE}
    & 1,792  & 1,536  
    & 2,560 &10,496\\
    \method{\model}
    & 56   & 56 & 56 & 57
    \\\bottomrule
        \end{tabular}
\end{table}

\section{Conclusions and Future Work}
In this paper, we explored multi-domain pre-training on graphs, with the objective of learning a comprehensive range of knowledge from diverse domains, and adapting to unseen domains in downstream applications. Our proposed approach \model\ leverages a series of domain tokens to unify the semantic spaces of multiple source domains, and extract domain-specific knowledge.
Moreover, we introduced dual prompts to align the target domain with pre-trained knowledge, including both unified multi-domain knowledge and a tailored mixture of source domain-specific knowledge. Finally, we conducted extensive experiments on six public datasets, demonstrating that \model\ significantly outperforms various state-of-the-art baselines.

In this work, we focused on text-free graphs, which are more challenging than text-attributed graphs due to the lack of textual descriptions. However, text-attributed graphs are also important, providing auxiliary knowledge and a gateway to connect with large language models. Therefore, a promising future direction is to integrate both text-free and text-attributed graphs across multiple domains for pre-training, in order to learn a more comprehensive range of knowledge.



\clearpage
\newpage
\bibliographystyle{ACM-Reference-Format}
\bibliography{references}
\clearpage
\newpage
\appendix
\section*{Appendices}
\renewcommand\thesubsection{\Alph{subsection}}
\renewcommand\thesubsubsection{\thesubsection.\arabic{subsection}}



\subsection{Further Descriptions of Datasets} \label{app.dataset}
In this section, we provide a comprehensive description of datasets.

(1) \emph{Cora}\footnote{\url{https://relational.fit.cvut.cz/dataset/CORA}} \cite{mccallum2000automating} includes 2,708 computing publications, each assigned to one of seven categories. The citation network has 5,429 links. Each publication in the dataset is depicted by a binary word vector, which shows the presence or absence of words from a dictionary containing 1,433 distinct words.

(2) \emph{Citeseer}\footnote{\url{https://nrvis.com/download/data/labeled/citeseer.zip}} \cite{sen2008collective} comprises 3,312 computing publications, each classified into one of six categories, which are different from those in \textit{Cora}. The citation network includes 4,732 links.  Each publication is described by a binary word vector that indicates the presence or absence of words from a dictionary containing 3,703 unique words.

(3) \emph{PubMed}\footnote{\url{https://github.com/kimiyoung/planetoid/raw/master/data}} \cite{sen2008collective} is composed of 19,717 biomedical publications from diabetes, each classified into one of three categories. The citation network comprises 44,338 links. Each publication in the dataset is depicted by a TF/IDF weighted word vector from a dictionary, indicating the presence of 500 unique words from the dictionary.

(4) \emph{Amazon Photo}\footnote{\url{https://github.com/shchur/gnn-benchmark/blob/master/data/npz/amazon_electronics_photo.npz}} \cite{shchur2018pitfalls}
 comprises 7,487 photography related products, categorized into one of eight categories. The co-purchasing network consists of 119,043 edges, which represent frequently bought-together product relationships. Each product is described by a feature vector derived from its metadata and reviews, and is labeled according to its category.

(5) \emph{Amazon Computers}\footnote{\url{https://github.com/shchur/gnn-benchmark/blob/master/data/npz/amazon_electronics_computers.npz}} \cite{shchur2018pitfalls} comprises 13,752 computer-related products, categorized into  ten groups. The co-purchasing network contains 245,861 edges, illustrating frequently bought-together product relationships. Each product is represented by a feature vector generated from its metadata and reviews, and is labeled according to its specific category.

(6) \emph{Reddit}\footnote{\url{https://data.dgl.ai/dataset/reddit.zip}} \cite{hamilton2017inductive} 
consists of 232,965 posts from the social media platform Reddit, categorized into 41 different subreddits. The interaction network includes 11,606,919 edges, representing interactions such as comments and replies between posts. Each post in the dataset is described by a feature vector derived from its textual content and metadata, consisting of 602 unique words. 

\subsection{Further Descriptions of Baselines} \label{app.baselines}
In this section, we summarize the datasets in Table~\ref{table.datasets} and present more details for the baselines used in our experiments.

\noindent (1) \textbf{End-to-end Graph Neural Networks}
\begin{itemize}
\item \textbf{GCN} \cite{kipf2016semi}: GCN utilizes a mean-pooling strategy for neighborhood aggregation to integrate information from neighboring nodes.
\item \textbf{GAT} \cite{velivckovic2017graph}: GAT also leverages neighborhood aggregation for end-to-end node representation learning, uniquely assigns varying attention weights to different neighbors, thereby adjusting their impact on the aggregation process.
\end{itemize}

\begin{table}[tbp]
\center
\caption{Summary of datasets. 
\label{table.datasets}}
\resizebox{0.85\linewidth}{!}{%
\begin{tabular}{@{}c|cccc@{}}
\toprule
	& \makecell[c]{Nodes} & \makecell[c]{Edges} & \makecell[c]{Feature\\dimension} & \makecell[c]{Node\\Classes}\\
\midrule
     Cora & 2,708 & 10,556 & 1,433 & 7 \\ 
     Citeseer & 3,327 & 9,104 & 3,703 & 6 \\ 
     Pubmed & 19,717 & 88,648 & 500 & 3 \\
     Photo & 7,650 & 238,162 & 745 & 8 \\
     Computers & 13,752 & 491,722 & 767 & 10 \\
     Reddit & 232,965& 114,615,892&602 & 41\\
 \bottomrule
\end{tabular}}
\end{table}

\noindent (2) \textbf{Graph Pre-training Models}
\begin{itemize}
\item \textbf{DGI} \cite{velivckovic2017graph}: DGI operates as a self-supervised pre-training methodology tailored for homogeneous graphs. It is predicated on the maximization of mutual information (MI), aiming to enhance the estimated MI between locally augmented instances and their global counterparts.
\item \textbf{InfoGraph} \cite{sun2019infograph}: Expanding upon DGI, InfoGraph is centered on graph-level tasks, endeavoring to maximize the alignment between node and graph embeddings.
\item \textbf{GraphCL} \cite{you2020graph}: GraphCL leverages a variety of graph augmentations for self-supervised learning, tapping into the intrinsic structural patterns of graphs. The overarching goal is to amplify the concordance between different augmentations throughout graph pre-training.
\end{itemize}

\noindent (3) \textbf{Graph Cross-domain Model}
\begin{itemize}
\item Hassani \cite{hassani2022cross}: Hassani proposes an attention-based graph encoder that utilizes contextual and topological views of graph to learn task-specific knowledge for rapid adaptation and task-agnostic knowledge for knowledge transfer.
\end{itemize}

\noindent (4) \textbf{Graph Prompt Models}
\begin{itemize}
\item \textbf{GPPT} \cite{sun2022gppt}: GPPT utilize a GNN model, pre-trained through a link prediction task. The downstream prompt module is specifically tailored for node classification task, which is unified as the pre-training link prediction task.
\item \textbf{GPF} \cite{fang2022universal}: GPF operates as a universal prompt-based tuning methodology tailored for pre-trained GNN models. It is predicated on the adaptation of the input graph's feature space, aiming to achieve an effect equivalent to any form of prompting function. 
\item \textbf{GraphPrompt} \cite{liu2023graphprompt}: GraphPrompt employs subgraph similarity calculation as a unified template to narrow the gap between pre-training and downstream tasks, inclusive of node and graph classification. A learnable prompt is subsequently tuned during the downstream adaptation to incorporate task-specific knowledge.

\end{itemize}

\noindent (5) \textbf{Multi-domain graph pre-training}
\begin{itemize}
\item \textbf{GCOPE} \cite{zhao2024all}: GCOPE is a multi-domain pre-training approach that cooperates graph datasets across multiple domains through a series of domain-specific interconnect virtual nodes that link nodes within the same domain. Its primary goal is to improve performance in downstream applications by leveraging the multiple domain knowledge in multiple source domains.
\end{itemize}

\subsection{Implementation Details} \label{app.parameters}

\stitle{General settings}\label{app.general-setting}
\noindent\textbf{Optimizer.} For all experiments, we use the Adam optimizer.

\noindent\textbf{Environment.} The environment in which we run experiments is:
\begin{itemize}[label=--]
    \item Linux version: 5.15.0-78-generic 
    \item Operating system: Ubuntu 18.04.5 LTS 
    \item CPU information: Intel(R) Xeon(R) Platinum 8352V 
    \item GPU information: GeForce RTX 4090 (24 GB)
\end{itemize}

\stitle{Further experiments setup.}\label{app.setup}
We conduct multi-domain pre-training with different source domains for various target domains, the correspondence is listed in Table~\ref{table.app.setting}.

\begin{table}[tbp]
\center
\caption{Settings of multi-domain transfer to unseen domains. 
\label{table.app.setting}}
\resizebox{0.95\linewidth}{!}{%
\begin{tabular}{@{}c|ccccc@{}}
\toprule
	Target domain & Cora & Citesser & Pubmed & Photo & Computers \\
\midrule
     Cora & $\times$ & $\checkmark$ & $\checkmark$ & $\checkmark$ & $\checkmark$ \\ 
     Citeseer & $\checkmark$ & $\times$ & $\checkmark$ & $\checkmark$ & $\checkmark$ \\ 
     Pubmed & $\checkmark$ & $\checkmark$ & $\times$ & $\checkmark$ & $\checkmark$ \\
     Photo & $\checkmark$ & $\checkmark$ & $\checkmark$ & $\times$ & $\checkmark$ \\
     Computers & $\checkmark$ & $\checkmark$ & $\checkmark$ & $\checkmark$ & $\times$ \\
     Reddit & $\checkmark$ & $\checkmark$ & $\checkmark$ & $\checkmark$ & $\checkmark$ \\
 \bottomrule
\end{tabular}}
\end{table}

To show the effectiveness of multi-domain pre-training, for \textit{Cora},\textit{ Citesser}, \textit{Pubmed}, \textit{Photos} and \textit{Computers}, we cooperate four datasets for multi-domain pre-training, while the rest for downstream adaptation. For further evaluate the performance of cross-domain adaptation, we leverage \textit{Cora},\textit{Citesser}, \textit{Pubmed}, \textit{Photos} and \textit{Computers} for multi-domain pre-training, while \textit{Reddit} for cross domain adaptation. Note that all pre-training methods are pre-trained on the above multi-domain datasets.

For downstream node and graph classification tasks, we conduct $m=1$ shot tasks. Additionally, we also vary the number of shots for $1 \le m\le 10$, to evaluate the robustness of \model.
We repeat the sampling 100 times to construct 100 $m$-shot tasks for both node and graph classification. 
For each task, we run with five different random seeds, leading to a total of 500 outcomes for node classification and grap classification tasks, and we report the average and standard deviation across these 500 results.

\stitle{Details of baselines.}
We utilize the officially provided code for all open-source baselines and reproduce the non-open-source GCOPE. Each model is tuned based on the settings recommended in their respective literature to achieve optimal performance.

For the baseline GCN \cite{kipf2016semi}, we employ a 3-layer architecture, and set the hidden dimensions to 256. 
For GAT \cite{velivckovic2017graph}, we employ a 2-layer architecture and set the hidden dimension to 64. Additionally, we apply 8 attention heads in the first GAT layer.

For DGI \cite{velivckovic2017graph}, we utilize a 1-layer GCN as the base model and set the hidden dimensions to 256. Additionally, we employ prelu as the activation function.
For InfoGraph \cite{sun2019infograph}, a 3-layer GCN is used as the base model, with its hidden dimensions set to 256.
For GraphCL \cite{you2020graph}, a 1-layer GCN is also employed as its base model, with the hidden dimensions set to 256. Specifically, we select edge dropping as the augmentations, with a default augmentation ratio of 0.2.

For Hassani \cite{hassani2022cross},We employ a 3-layer GCN is used as the base model for all datasets, with the hidden dimensions set to 256.

For GPPT \cite{sun2022gppt}, we utilize a 2-layer GraphSAGE as its base model, setting the hidden dimensions to 256. For base GraphSAGE, we also employ a mean aggregator.
For GraphPrompt \cite{liu2023graphprompt}, a 3-layer GCN is used as the base model for all datasets, with the hidden dimensions set to 256.
For GPF \cite{fang2022universal}, employs a 5-layer GCN as the base model for all datasets, following the recommended settings. The hidden dimensions are set to 256.

For GCOPE \cite{zhao2024all}, we employ a 2-layer GCN as the base model and set the hidden dimensions to 100. Downstream adaptation is achieved through fine-tuning, as it is reported to yield the best performance in their literature.

For all baselines except for GCOPE, we set the unified feature dimensions to 50, the same as our \model. For GCOPE, we adhere to the recommended settings and set the unified feature dimensions to 100.

\stitle{Details of \model.}
For our proposed \model, we utilize a 3-layer GCN as the base model for all datasets, with the hidden dimensions set to 256. We set the unified feature dimensions to 50.

\end{document}